\documentclass[default]{sn-jnl}

\usepackage{graphicx}%
\usepackage{multirow}%
\usepackage{amsmath,amssymb,amsfonts}%
\usepackage{amsthm}%
\usepackage{mathrsfs}%
\usepackage[title]{appendix}%
\usepackage{xcolor}%
\usepackage{textcomp}%
\usepackage{manyfoot}%
\usepackage{booktabs}%
\usepackage{algorithm}%
\usepackage{algorithmicx}%
\usepackage{algpseudocode}%
\usepackage{listings}%
\usepackage{comment}

\begin{document}

\title{MuMUR : \underline{Mu}ltilingual \underline{M}ultimodal \underline{U}niversal \underline{R}etrieval}


\author*[1]{\fnm{Avinash Madasu} }\email{avinash.madasu@intel.com}

\author[1]{\fnm{Estelle Aflalo} }\email{estelle.aflalo@intel.com}

\author[1]{\fnm{Gabriela Ben Melech Stan}}\email{gabriela.ben.melech.stan@intel.com}

\author[1]{\fnm{Shachar Rosenman}}\email{shachar.rosenman@intel.com}

\author[1]{\fnm{Shao-Yen Tseng} }\email{shao-yen.tseng@intel.com}

\author[2]{\fnm{Gedas Bertasius} }\email{gedas@cs.unc.edu}

\author[1]{\fnm{Vasudev Lal} }\email{vasudev.lal@intel.com}

\affil[1]{\orgdiv{Cognitive Computing Research}, \orgname{Intel Labs}}

\affil[2]{\orgdiv{Department of Computer Science}, \orgname{University of North Carolina at Chapel Hill}}


\abstract{Multi-modal retrieval has seen tremendous progress with the development of vision-language models. However, further improving these models require additional labelled data which is a huge manual effort. In this paper, we propose a framework MuMUR, that utilizes  knowledge transfer from a multilingual model to boost the performance of multi-modal (image and video) retrieval.  We first use state-of-the-art machine translation models to construct pseudo ground-truth multilingual visual-text pairs. We then use this data to learn a joint vision-text representation where English and non-English text queries are represented in a common embedding space based on pretrained multilingual models. We evaluate our proposed approach on a diverse set of retrieval datasets: five video retrieval datasets such as MSRVTT, MSVD, DiDeMo, Charades and MSRVTT multilingual, two image retrieval datasets such as Flickr30k and Multi30k . Experimental results demonstrate that our approach achieves state-of-the-art results on all video retrieval datasets outperforming previous models. Additionally, our framework MuMUR significantly beats other multilingual video retrieval dataset. We also observe that MuMUR exhibits strong performance on image retrieval. This demonstrates the universal ability of MuMUR to perform retrieval across all visual inputs (image and video) and text inputs (monolingual and multilingual).}

\maketitle

\section{Introduction}\label{sec1}


The task of multimodal retrieval aims to retrieve images or videos that are semantically similar to a given text query and vice-versa. In the world of computer vision, image retrieval and video retrieval have been treated as separate tasks. For example, transformer models pretrained on large amounts of image-text pairs are then fine tuned for the task of image retrieval ~\cite{chen2019uniter, jia2021scaling, li2022blip, li2021align, li2022align}. In contrast, video retrieval models are developed in two parallel directions. The first line of work \cite{bain2021frozen,li2022align,ge2022bridging} focused on video-text pretraining on large-scale datasets like Howto100M \cite{miech2019howto100m} and WebVid-2M \cite{bain2021frozen}. While the second line of work  \cite{luo2021clip4clip} focused on using pretrained image features like CLIP \cite{radford2021learning} for video retrieval often surpassing the models pretrained on video datasets. 

\begin{figure}[!t]
  \centering
  \includegraphics[width=0.7\columnwidth]{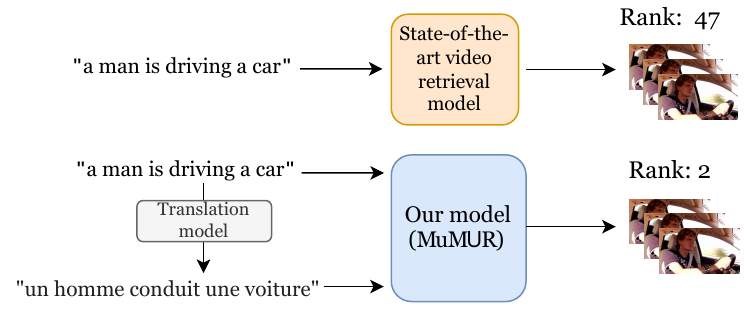}
    \caption{Illustration of the improved video retrieval ranking for a test sample in MSRVTT dataset. For the current state-of-the-art video retrieval model, the rank of the ground truth video is 47. When multilingual data is used as knowledge transfer, the ranking of the ground truth video improved significantly from 47 to 2.
  }

  \label{problem_def}
\end{figure}

Most of the current multimodal retrieval datasets typically contain around 10k visual inputs and the corresponding captions of maximum lengths ranging from 30 to 60. In multimodal retrieval, the text encoder projects the input text caption and visual inputs (image or video) into a common embedding space. With longer captions, the text embeddings might lose required contextual information resulting in incorrect retrievals. One could address this by incorporating more structured knowledge (e.g., parts of speech, dependency graphs) in the text encoder \cite{cao2022visual}. However, a drop in performance is observed \cite{cao2022visual} with the addition of structural knowledge in a text-to-video retrieval setting. One could argue that the reason might be the smaller retrieval datasets and creating meaningful structural knowledge becomes a challenging task.

There are over 6000 languages in the world each having its own vocabulary of words, grammar and morphology. However, there exists an overlap of knowledge among these languages ~\cite{hu2020xtreme, artetxecross, wu2019surprising}.
In Natural Language Processing, some works \cite{conneau2019cross} explored this idea of using multilingual data to improve the performance on monolingual English datasets. Conneau et al. introduced a new pretraining objective: Translation Language Modeling (TLM) in which random words were masked in the concatenated sentences of English and multilingual data and the model predicts the masked words. There, the objective was to use multilingual context to predict masked English words if the English context was not sufficient and vice-versa.

Most of the current works in multimodal retrieval are primarily focused on monolingual English datasets. Recently, few works have proposed models that can perform retrieval in a multilingual setting ~\cite{huang2021multilingual, zhou2021uc2, ni2021m3p, burns2020learning}.
However, these works use large amounts of multilingual data for pretraining and are tuned for retrieval on each language. Moreover, these models are also limited to performing either image retrieval or video retrieval. In addition, these models show a drop in performance on English retrieval datasets even though there is a boost in performance on multilingual data.

In this work, we are interested in two objectives \textbf{(i) design a model capable of performing retrieval on both images and videos (ii) design a model capable of performing retrieval on multiple languages including English}. Multilingual data serves as a powerful knowledge augmentation for monolingual models \cite{conneau2019cross}. Nevertheless, creating multilingual data requires huge human effort. To overcome this, we use state-of-the-art machine translation models \cite{tang2020multilingual} to convert English text captions into other languages. Specifically, we choose languages whose  performance on XNLI benchmark \cite{conneau2018xnli} is comparable to that of English (i.e., French, German, Spanish). With this, we create high quality multilingual data without requiring human labelling. To the best of our knowledge, this is the first work that uses multilingual knowledge transfer to improve multimodal retrieval.

We propose a new framework MuMUR : \underline{Mu}ltilingual \underline{M}ultimodal \underline{U}niversal \underline{R}etrieval that is capable of performing both image and video retrieval on multilingual datasets. This framework is based on CLIP \cite{radford2021learning} to effectively utilize and adapt the multilingual knowledge transfer. Our model takes a visual input, English text caption and multilingual text caption as inputs and extracts joint visual-text representations. The multilingual text representations should act as a knowledge augmentation to the English text representations aiding in retrieval. For this purpose, we introduce a dual cross-modal (DCM) encoder block which learns the similarity between English text representations and visual representations. In addition, the DCM encoder block also associates the visual representations with the multilingual text representations. In the common embedding space, our model learns the important contextual information from multilingual representations which is otherwise missing from the English text representations effectively serving as  knowledge transfer. 

We validate our proposed model on a comprehensive set of image retrieval datasets: Flickr30k ~\cite{plummer2015flickr30k} and video retrieval datasets: MSRVTT-9k \cite{xu2016msr}, MSRVTT-7k \cite{xu2016msr}, MSVD \cite{chen2011collecting}, DiDeMo \cite{anne2017localizing} and Charades \cite{sigurdsson2016hollywood}.  We show that our approach achieves state-of-the-art results, outperforming previous models on most datasets. In addition to the evaluation on monolingual retrieval datasets, we also compare the performance of our model on multilingual datasets Multi30k ~\cite{elliott2016multi30k} and MSRVTT multilingual.\cite{huang2021multilingual}. Experimental results demonstrate that MuMUR achieves state-of-the-art results on all English video retrieval datasets and significantly outperforms previous models on multilingual video retrieval in a zero-shot setting. Furthermore, our model MuMUR establishes new benchmark on multilingual image retrieval while achieving strong performance on English image retrieval datasets. These results demonstrate the universal capability of MuMUR to perform all types of multimodal and multilingual retrieval.

To summarize, our contributions are as follows: (i) We generate multilingual data using external state-of-the-art machine translation models. (ii) We propose a model that is capable of knowledge transfer from multilingual data to improve the performance of multimodal retrieval. (iii) We evaluate the proposed framework on six English retrieval benchmarks and achieve state-of-the-art results in both text-to-visual and visual-to-text retrieval settings. (iv) Finally, we demonstrate that our model significantly outperforms previous approaches on multilingual retrieval datasets.

\section{Related work}
\subsection{Multimodal retrieval}
Pre-train and then fine-tune is the most popular paradigm involving image retrieval ~\cite{chen2019uniter, li2020hero, li2021align, li2022blip}. These models are pre-trained on huge amounts of image text pairs such as Conceptual Captions ~\cite{changpinyo2021conceptual}, Visual Genome ~\cite{krishna2017visual} and SBU ~\cite{lu202012} and tested on image retrieval datasets such as Flickr30k ~\cite{plummer2015flickr30k} and COCO ~\cite{karpathy2015deep}.

The task of video retrieval has seen tremendous progress in the recent years. This is partly due to the availability of large-scale video datasets like HowTo100M \cite{miech2019howto100m} and WebVid-2M \cite{bain2021frozen}. Besides the adaption of transformers to image tasks like image classification \cite{dosovitskiy2020image} spurred the development of models based on transformers. However, videos require  computationally more memory and compute power and can be infeasible to compute self-attention matrices. With the introduction of more efficient architectures \cite{bertasius2021space} large-scale pretraining on videos became a possibility. In this direction, several transformer based architectures \cite{bain2021frozen,ge2022bridging,madasu2022learning,madasu2023improving} were proposed and pretrained on large video datasets which achieved state-of-the-art results on downstream video retrieval datasets in both zero-shot and fine-tuning settings. 

In a parallel direction, a few works \cite{luo2021clip4clip} have adopted image level features pretrained on large scale image-text pairs to perform video retrieval. Surprisingly, these works have performed significantly better than the models that are pretrained from scratch on large scale video datasets. Compared to these models, our approach completely differs in the architecture and the training methodology. 

\subsection{Multilingual training}
The recent success of multimodal image-text models on a variety of tasks, such as retrieval and question-answering, has been mostly limited to monolingual models trained on English text. This is mainly due to the availability and high-quality of English-based multimodal datasets. Recent work indicates that incorporating a second language or a multilingual encoder, thus creating a shared multilingual token embedding space, can improve monolingual pure-NLP downstream tasks \cite{conneau2019cross}. This concept was rapidly embraced for training multimodal models. Previous works had used images as a bridge for translating between two languages, without using a language-to-language shared dataset for training \cite{chen2011collecting,suris2022globetrotter,sigurdsson2020visual}. 

Recent work has focused on multimodal tasks, such as image retrieval, aiming to add multilingual capabilities to multimodal models \cite{burns2020learning,gella2017image}. The work often indicates that incorporating a second language during training of multimodal models, improves performance on single-language multimodal tasks such as image retrieval, compared to multimodal models that were trained on a single language \cite{gella2017image,kim2020mule,wehrmann2019language}. MULE \cite{kim2020mule}, which is a multilingual universal language encoder trained on image-multilingual text pairs, showed an improvement on image-sentence retrieval tasks of up to 20\% compared to monolingual models. Nevertheless, all these previous works focus on designing models separately for image and video retrieval. Our objective is to use multilingual knowledge transfer to improve the performance on current image and video retrieval datasets. In addition, our model is capable of performing multilingual retrieval on more than 10 languages.

\begin{figure*}[!t]
  \centering
  \includegraphics[width=0.8\textwidth]{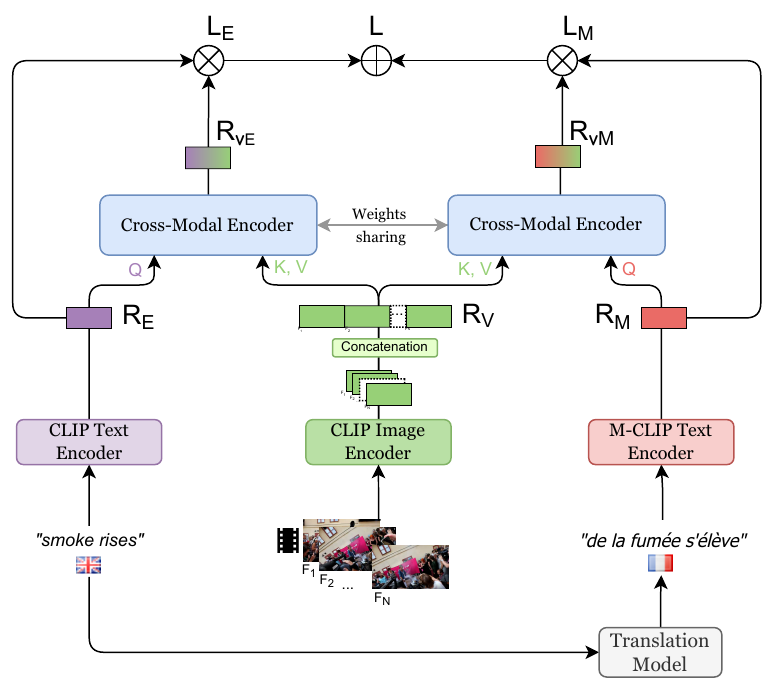}
  \caption{\textbf{Illustration of the proposed MuMUR model}. The model takes as visual input (image or video), a corresponding English text query and a translated multilingual query. The multilingual text query is obtained using the off-the-shelf machine translation model. It is used only for the inference and is not part of the architecture. The video and English text features are extracted using CLIP model whereas multilingual text features are extracted using M-CLIP model. The features are then passed onto a cross-model encoder to learn the association in a common embedding space. Cross-entropy loss is then applied to measure the similarity between text features $R_{E}$ and $R_{vE}$, $R_{M}$ and $R_{vM}$. The final loss is the sum of both the losses.}
  \label{MuMUR_figure}
\end{figure*}

\section{MuMUR : Multilingual Multimodal Universal Retrieval}
In this section, we introduce our framework MuMUR : Multilingual Multimodal Universal Retrieval. We first describe the problem statement, then the multilingual data augmentation strategy and finally go over the proposed approach that enables knowledge transfer from multilingual data for video retrieval.

\subsection{Problem statement}
Given a set of visual data $V$, their corresponding English text captions $E$ and related multilingual text captions $M$, our goal is to learn similarity functions $s_{1}(v_{i}, e_{i})$ and $s_{2}(v_{i}, m_{i})$ ($v_{i} \in V$, $e_{i} \in E$ and $m_{i} \in M$). In other words, we propose a framework MuMUR that enables end-to-end learning on a tuple of visual input, English text caption and multilingual text caption by bringing closer the joint representations of those three elements. Specifically, for each visual input $V$ and the English text captions $E$, we generate the multilingual translations $M$ using external state-of-the-art machine translation models \cite{tang2020multilingual}. Next, we present the proposed approach that facilitates the end-to-end learning using multilingual data.

\subsection{Approach}
Our model, illustrated in  Figure \ref{MuMUR_figure}, is comprised of three components: (i) visual encoder (ii)  text encoder  (iii) dual cross modal encoder. Next, we describe the framework in detail. 
\subsubsection{Visual Encoder}
Given a visual input $V$, we consider uniformly sampled clips $C \in R^{N_{v} \times H \times W \times 3}$ where $N_{v}$ is the number of frames (1 for images), $H$ and $W$ are the spatial dimensions of a RGB frame. We then use a pretrained CLIP-ViT image encoder  \cite{radford2021learning} to extract the frame embeddings $F_{v} \in R^{N_{v} \times D_{v}}$ where $D_{v}$ denotes the dimensions of the frame embeddings. The frame embeddings are concatenated to obtain the final representation for the video $V$.
\subsubsection{Text Encoder}
Let the inputs English text caption be $E$ and multilingual text caption $M$ of lengths $p$ and $q$ respectively. We use a pretrained CLIP-ViT text encoder to convert the English text caption into a sequence of embeddings $R_{E} = R^{E_{p} \times D_{E}}$ where $D_{E}$ denote the embedding dimensions. We consider the representation of the token $[EOS]$ as the final representation of English text caption. To encode multilingual text caption $M$, we use a M-CLIP\footnote{https://github.com/FreddeFrallan/Multilingual-CLIP} model which is a multilingual clip model pretrained on multilingual text and image pairs. Specifically, the multilingual text caption is converted into a sequence of embeddings $R_{M} = R^{M_{q} \times D_{M}}$ where $D_{M}$ denote the embedding dimensions. Similar to the CLIP model, we consider the $[EOS]$ representation as the final representation of M-CLIP model.
\subsubsection{Dual cross-modal encoder (DCM)}
Our goal is to closely associate the visual embeddings $R_{v}$, English text embeddings $R_{E}$ and multilingual text embeddings $R_{M}$ in a common embedding space. For this purpose, we propose a dual cross-modal encoder (DCM). To incorporate textual information into visual features and to learn visual features that are semantically most similar to text features, we use multi-head attention. The text features are used as the queries whereas the visual features are used as the keys and values. 

\begin{equation}
    r_{vE} = Attention(T_{E}, F_{v}, F_{v})
\end{equation}
\begin{equation}
    r_{vM} = Attention(M_{E}, F_{v}, F_{v})
\end{equation}
where multi-head attention ($Attention$) is defined as:
\begin{equation}
    Attention(Q, K, V) = Softmax(\frac{QK^{T}}{\sqrt{d}})V
\end{equation}
Here $Q$, $K$ and $V$ are same as the original multi-head attention matrices in the transformer encoder.
We then apply a fully connected layer on the attention outputs and finally layer normalization to obtain $R_{vE}$ and $R_{vM}$.
\begin{equation}
    R_{vE} = LN(FC(r_{vE}) + r_{vE})
\end{equation}
\begin{equation}
    R_{vM} = LN(FC(r_{vM}) + r_{vM})
\end{equation}
where $FC$ is the fully connected layer and $LN$ is the layer normalization layer. 
\subsubsection{Loss}
We use the standard image-text or video-text matching loss \cite{zhaiclassification} to train the model. It is measured as the dot product similarity between matching text embeddings and visual embeddings in a batch. 
First, we compute the loss $L_{E}$ between $R_{vE}$ and $R_{E}$ and then compute the loss $L_{M}$ between $R_{vM}$ and $R_{M}$. The final loss is the sum of losses $L_{E}$ and $L_{M}$.
\begin{equation}
    L = L_{E} + L_{M}.
\end{equation}
where
$L_{E}$ = $L_{E}^{t2v}$ + $L_{E}^{v2t}$ and $L_{M}$ = $L_{M}^{t2v}$ + $L_{M}^{v2t}$
\begin{equation}
\begin{aligned}
\mathcal{L}_{E}^\mathrm{v2t} & =
-\frac{1}{B}\sum_{i=1}^B\log\frac{ \exp({R_{E}}^{(i)} \cdot     R_{vE}^{(i)})}{\sum_{j=1}^{B}\exp({R_{E}}^{(i)}\cdot R_{vE}^{(j)})},\\
\end{aligned}
\end{equation}

\begin{equation}
\begin{aligned}
\mathcal{L}_{E}^\mathrm{t2v} & = 
-\frac{1}{B}\sum_{i=1}^B\log\frac{ \exp({R_{vE}}^{(i)}\cdot  R_{E}^{(i)})}{\sum_{j=1}^{B}\exp({R_{vE}}^{(i)}\cdot R_{E}^{(j)})}.\\
\end{aligned}
\end{equation}

\begin{equation}
\begin{aligned}
\mathcal{L}_{M}^\mathrm{v2t} & =
-\frac{1}{B}\sum_{i=1}^B\log\frac{ \exp({R_{M}}^{(i)} \cdot     R_{vM}^{(i)})}{\sum_{j=1}^{B}\exp({R_{M}}^{(i)}\cdot R_{vM}^{(j)})},\\
\end{aligned}
\end{equation}

\begin{equation}
\begin{aligned}
\mathcal{L}_{M}^\mathrm{t2v} & = 
-\frac{1}{B}\sum_{i=1}^B\log\frac{ \exp({R_{vM}}^{(i)}\cdot  R_{M}^{(i)})}{\sum_{j=1}^{B}\exp({R_{vM}}^{(i)}\cdot R_{M}^{(j)})}.\\
\end{aligned}
\end{equation}
\subsubsection{Inference}
During inference for English retrieval datasets, we freeze the multilingual text encoder and measure the retrieval performance only using $R_{E}$ and $R_{vE}$. Similarly, for multilingual datasets, we freeze the English text encoder and calculate the retrieval score using $R_{M}$ and $R_{vM}$. 
\section{Experiments}
\subsection{Datasets}
Our goal is to design a universal multimodal multilingual retrieval model. Therefore, our experiments focus on evaluating on both image and video retrieval datasets comprising of monolingual and multilingual captions.
\subsubsection{Video Retrieval}
We perform experiments on six standard text-video retrieval datasets: MSRVTT-9k and MSRVTT-7k splits \cite{xu2016msr}, MSVD \cite{chen2011collecting}, DiDeMo \cite{anne2017localizing}, Charades \cite{sigurdsson2016hollywood} and MSRVTT multilingual \cite{huang2021multilingual}.

\textbf{MSRVTT} contains 10K videos with each video ranging from 10 to 32 seconds and 200K captions. We report the results both on MSRVTT-9k and MSRVTT-7k datasets following \cite{luo2021clip4clip}.

\textbf{MSVD} consists of 1970 videos and 80K descriptions. We use the standard training, validation and testing splits following \cite{luo2021clip4clip}. In this dataset, each video has multiple captions and are treated as independent samples during testing.

\textbf{DiDeMo} is made up of 10K videos and 40K localized descriptions of the videos. We concatenate all the sentences for each video and evaluate the paragraph-to-video retrieval following \cite{lin2022eclipse,luo2021clip4clip}.

\textbf{Charades} contains of 9848 videos and each video is associated with a caption. We use the standard training and test splits following \cite{lin2022eclipse}.

\textbf{MSRVTT multilingual} is a multilingual version of MSRVTT in which the English captions are translated into nine different languages. We use the standard splits following \cite{huang2021multilingual}. 

\subsubsection{Image Retrieval}
We evaluate the proposed approach MuMUR on the following image retrieval datasets:

\textbf{Flickr30k} contains 31000 images with each image containing single caption for training and validation data and 5 captions for testing data. We follow the standard splits of 29k/1k/1k ~\cite{li2022blip}.

\textbf{Multi30K} ~\cite{elliott2016multi30k} is a multilingual version
of Flickr30k ~\cite{plummer2015flickr30k} in which the English text captions are translated into German (de), French (fr) and Czech (cs) languages. We use the training, validation and testing splits of 29k/1k/1k following ~\cite{ni2021m3p}.
\subsection{Metrics}
For evaluating the performance of models, we use recall at rank $K$ ($R@1$, $R@5$, $R@10$), median rank (MedR) and mean rank (MnR). Unless specified, the values reported are the mean of three runs with different seeds. 

\begin{table*}[ht]
	\begin{center}
		\resizebox{\textwidth}{!}
		{
		\begin{tabular}{c|c|ccccc|ccccc} %
	    \hline
        \multicolumn{1}{c}{} & \multicolumn{1}{c}{} & \multicolumn{5}{c}{Text-to-Video Retrieval} &\multicolumn{5}{c}{Video-to-Text Retrieval} \\
        \hline
	    Type & Model & R@1  & R@5  & R@10  & MdR & MnR & R@1 & R@5  & R@10 & MdR & MnR \\
        \hline
	    \multicolumn{1}{c|}{\multirow{ 13}{*}{\rotatebox{90}{Others}}} 
	    & JsFusion \cite{yu2018joint}             & 10.2 & 31.2 & 43.2 & 13.0 & -    & -    &    - & -    &   - & - \\
	    & HT \cite{miech2019howto100m} & 14.9 & 40.2 & 52.8 & 9.0  & -    & -    & -   & -    &   - & - \\
	    & HERO \cite{li2020hero}                    & 20.5 & 46.8 & 60.9 & -  & - & - & - & - & - & - \\
	    & CE \cite{liu2019use}                    & 20.9 & 48.8 & 62.4 & 6.0  & 28.2 & 20.6 & 50.3 & 64.0 & 5.3 & 25.1 \\
	    & ClipBERT \cite{lei2021less}             & 22.0 & 46.8 & 59.9 & - & - & - & - & - & - & - \\
	    & SupportSET \cite{patrick2020support}                & 27.4 & 56.3 & 67.7 & 3.0  & -    & -    & -    & -    & -   & - \\
	    & VideoCLIP \cite{xu2021videoclip}                & 30.9 & 55.4 & 66.8 & 4.0  & -    & -    & -    & -    & -   & - \\
	    & FrozenInTime \cite{bain2021frozen}                & 31 & 59.5 & 70.5 & 3.0  & -    & -    & -    & -    & -   & - \\
	    & CLIP \cite{radford2021learning}                & 31.2 & 53.7 & 2.6 & 4.0  & -    & -    & -    & -    & -   & - \\
	    & HIT \cite{liu2021hit}        & 30.7 & 60.9 & 73.2 & 2.6  & -    & 32.1 & 62.7 & 74.1 & 3.0 & - \\
	    & AlignPrompt \cite{li2022align}   & 33.9 & 60.7 & 73.2 & -  & -    & -    & -   & -    &   - & - \\
	    & All-in-one \cite{wang2022all}   & 34.4 & 65.4 & 75.8 & -  & -    & -    & -   & -    &   - & - \\
	    & MDMMT \cite{dzabraev2021mdmmt}            & 38.9 & 69.0 & 79.7 & \textbf{2.0}  & -    & -    &    - & -    &   - & - \\
	    \hline
	    \multicolumn{1}{c|}{\multirow{ 7}{*}{\rotatebox[origin=c]{90}{CLIP based}}} & CLIP4Clip \cite{luo2021clip4clip} & 44.5 & 71.4 & 81.6	& -  & 15.3 & 43.1 & 70.5 & 81.2 & \textbf{2.0} & 12.4 \\	
        & VCM \cite{cao2022visual} & 43.8 & 71.0 & 80.9	& \textbf{2.0}  & 14.3 & 45.1 & 72.3 & 82.3 & \textbf{2.0} & 10.7 \\
	    & MCQ \cite{ge2022bridging} & 44.9 & 71.9 & 80.3 & \textbf{2.0} & 15.3  & -    & -    & -   & -    &   - \\
	    & MILES \cite{ge2022miles} & 44.3 & 71.1 & 80.8 & \textbf{2.0} & 14.7 & -    & -    & -   & -    &   - \\
	    & CAMoE \cite{cheng2021improving} & 44.6 & \textbf{72.6} & 81.8 & \textbf{2.0} & \textbf{13.3} & 45.1 & 72.4 & 83.1 & \textbf{2.0} & 10.0 \\
	    & CLIP2Video \cite{fang2021clip2video} & 45.6 & \textbf{72.6} & 81.7 & \textbf{2.0} & 14.6 & 43.5 & 72.3 & 82.1 & \textbf{2.0} & 10.2 \\
	    & CLIP2TV \cite{gao2021clip2tv} & 46.1 & 72.5 & \textbf{82.9} & \textbf{2.0} & 15.2 & 43.9 & 70.9 & 82.2 & \textbf{2.0} & 12.0 \\
	    \hline 
	   \multicolumn{1}{c|}{\rotatebox[origin=c]{0}{Ours}} & \textbf{MuMUR} & \textbf{46.6} & \textbf{72.6} & 82.2 & \textbf{2.0} & 13.9 & \textbf{45.5} & \textbf{73.4} & \textbf{84.7} & \textbf{2.0} & \textbf{8.07} \\
	   \hline
		\end{tabular}}
		\caption{Text-to-video and video-to-text retrieval results on MSR-VTT dataset 9k split. Recall at rank $1$ (R@1)$\uparrow$, rank $5$ (R@5)$\uparrow$, rank $10$ (R@10)$\uparrow$, Median Rank (MdR)$\downarrow$ and Mean Rank (MnR)$\downarrow$ are reported. Results of other methods taken from mentioned references. Our model surpasses previous state-of-the-art performance. In video-to-text retrieval, our model achieved 1.6 points boost in performance.}
        \label{tab: MSRVTT_9k}
	\end{center}
\end{table*}

\subsection{Hardware}
Training was conducted on compute nodes that consist of Intel Xeon processors equipped with Intel\textregistered Gaudi\textregistered2 AI accelerators with 96GB of vRAM

\subsection{Implementation Details}
We use translations of French for the multilingual inputs to train the MuMUR model. 
The visual encoder and the English text encoder are initialized with CLIP-ViT-B-32. The multilingual text encoder is initialized with M-CLIP-ViT-B-32. The dimension size of the video, English caption and multilingual caption representations is 512. The dual cross-model encoder is initialized randomly and trained from scratch. The dimension size of the key, query and value projection layers is 512. The fully connected layer in the transformer has a size of 512 and a dropout of 0.4 is applied on this layer. We use 16 frames for MSRVTT-9k, MSRVTT-7k and MSVD datasets, 42 frames for DiDeMo and Charades datasets. The maximum sequence length is set to 32 for MSRVTT-9k and MSRVTT-7k, 64 for DiDeMo and 30 for charades dataset. The model is trained using AdamW \cite{loshchilov2018decoupled} a learning rate of 1e-4 and a cosine decay of 1e-6. The MSRVTT-9k and MSRVTT-7k datasets are trained with a batch size of 32 and for 15 epochs. The MSVD dataset is trained with a batch size of 32 and for 5 epochs. The DiDeMo and charades datasets are trained with a batch size of 16 for 12 and 15 epochs respectively.

\begin{table}[ht]
	\begin{center}
		\resizebox{0.8\textwidth}{!}
		{
		\begin{tabular}{c|cccc} %
        \hline
	    Model & R@1 ($\uparrow$) & R@5 ($\uparrow$) & R@10($\uparrow$) & MdR ($\downarrow$) \\
        \hline
	    HowTo100M \cite{miech2019howto100m}             & 10.2 & 31.2 & 43.2 & 13.0  \\
	    ActBERT \cite{zhu2020actbert}             & 8.6 & 23.4 & 33.1 & 36.0  \\
	    NoiseE \cite{amrani2021noise}             & 17.4 & 41.6 & 53.6 & 8.0  \\
	    ClipBERT \cite{lei2021less}             & 22.0 & 46.8 & 59.9 & 6.0  \\
	    CLIP4clip- \cite{luo2021clip4clip}             & 42.1 & 71.9 & 81.4 & \textbf{2.0}  \\
	    Singularity \cite{lei2022revealing}             & 42.7 & 69.5 & 78.1 & \textbf{2.0}  \\
	    \hline
	    \textbf{MuMUR}             & \textbf{44.8} & \textbf{72.0} & \textbf{82.5} & \textbf{2.0}  \\
	    \hline
		\end{tabular}}
		\caption{Text-to-video retrieval results on MSR-VTT - 7k split. Recall at rank-$1$ (R@1), rank-$5$ (R@5), rank-$10$ (R@10), Median Rank (MdR) are reported. Results of other methods taken from mentioned references.}
		\label{tab: MSRVTT_7k}
	\end{center}
\end{table}

\begin{table}[ht]
	\begin{center}
		\resizebox{0.8\textwidth}{!}
		{
		\begin{tabular}{c|cccc} %
        \hline
	    Model & R@1 ($\uparrow$) & R@5 ($\uparrow$) & R@10($\uparrow$) & MdR ($\downarrow$) \\
        \hline
	    VSE \cite{fang2021clip2video}             & 12.3& 30.1& 42.3 &14.0  \\
	    CE \cite{liu2019use}             & 19.8 &49.0& 63.8& 6.0  \\
	    SSML \cite{amrani2021noise}             & 20.3 & 49.0 & 63.3 & 6.0  \\
	    SUPPORT-SET \cite{patrick2020support}             & 28.4 & 60.0 & 72.9 & 4.0  \\
	    FROZEN \cite{bain2021frozen}             & 33.7 & 64.7 & 76.3 & 3.0  \\
	    CLIP \cite{radford2021learning} & 37.0 & 64.1 & 73.8 & 3.0 \\
	    CLIP4Clip \cite{luo2021clip4clip}             & 46.2 & 76.1 & 84.6 & 2.0  \\
	    CLIP2Video \cite{fang2021clip2video}             & 47.0 & 76.8 & 85.9 & 2.0  \\
	    \hline
	    \textbf{MuMUR}             & \textbf{47.2} & \textbf{77.3} & \textbf{86.2} & \textbf{2.0}  \\
	    \hline
		\end{tabular}}
		\caption{Text-to-video retrieval results on MSVD dataset (multi-caption evaluation). Recall at rank-$1$ (R@1), rank-$5$ (R@5), rank-$10$ (R@10), Median Rank (MdR) are reported. Results of other methods taken from mentioned references.}
		\label{tab: MSVD}
	\end{center}
\end{table}

\begin{table}[ht]
	\begin{center}
		\resizebox{0.9\textwidth}{!}
		{
		\begin{tabular}{c|cccc} %
        \hline
	    Model & R@1 ($\uparrow$) & R@5 ($\uparrow$) & R@10 ($\uparrow$) & MdR ($\downarrow$) \\
        \hline
	    S2VT \cite{venugopalan2015sequence}             & 11.9 & 33.6 & - & 13.0  \\
	    FSE \cite{zhang2018cross}             & 13.9 & 36 & - & 11.0   \\
	    CE \cite{liu2019use}             & 16.1 & 41.1 & - & 8.3  \\
	    ClipBERT \cite{lei2021less}             & 20.4 & 48.0 & 60.8 & 6.0  \\
	    FrozenInTime \cite{bain2021frozen}             & 31.0 & 59.8 & 72.4 & 3.0   \\
	    OA-Trans \cite{wang2022object}             & 34.8 & 64.4 & 75.1 & 3.0  \\
	    CLIP4clip \cite{luo2021clip4clip}             & 43.4 & 70.2 & 80.6 & \textbf{2.0}  \\
	    CLIP2TV \cite{gao2021clip2tv}             & 43.9 & 70.5 & 79.8 & \textbf{2.0}   \\
	    TS2-Net \cite{liu2022ts2}             & 41.8 & 71.6 & 82.0 & \textbf{2.0}   \\
	    ECLIPSE \cite{lin2022eclipse}             & 44.2 & 70.0 & 80.2 & \textbf{2.0}   \\
	    \hline
	    \textbf{MuMUR}            & \textbf{44.4} & \textbf{74.3} & \textbf{83.1} & \textbf{2.0}  \\
	    \hline
		\end{tabular}}
		\caption{Text-to-video retrieval result on DiDeMo dataset. Recall at rank-$1$ (R@1), rank-$5$ (R@5), rank-$10$ (R@10), Median Rank (MdR) are reported. Results of other methods taken from mentioned references.}
		\label{tab: DiDeMo}
	\end{center}
\end{table}

\begin{table}[!h]
	\begin{center}
		\resizebox{0.9\textwidth}{!}
		{
		\begin{tabular}{c|ccccc} %
        \hline
	    Model & R@1 ($\uparrow$) & R@5 ($\uparrow$) & R@10 ($\uparrow$) & MdR & MnR \\
        \hline
	    ClipBERT \cite{lei2021less}             & 6.7 & 17.3 & 25.2 & 32.0 & 149.7  \\
	    FrozenInTime \cite{bain2021frozen}            & 11.9 & 28.3 & 35.1 & 17.0 & 103.8  \\
	    CLIP4clip \cite{luo2021clip4clip}             & 13.9 & 30.4 & 37.1 & 14.0 & 98.0 \\
	    ECLIPSE \cite{lin2022eclipse}            & 15.7 & 32.9 & 42.4 & 16.0 & 84.9  \\
	    \hline
	    \textbf{MuMUR}           & \textbf{16.6} & \textbf{37.5} & \textbf{50.0} & \textbf{10.0} & \textbf{52.7} \\
	    \hline
		\end{tabular}}
		\caption{Text-to-video retrieval result on charades dataset. Recall at rank-$1$ (R@1), rank-$5$ (R@5), rank-$10$ (R@10), Median Rank (MdR) are reported. Results reported are taken from \cite{lin2022eclipse}.}
		\label{tab: Charades}
	\end{center}
\end{table}

\begin{table*}[ht]
	\begin{center}
		\resizebox{\textwidth}{!}
		{
		\begin{tabular}{c|ccc|ccc} %
	    \hline
         \multicolumn{1}{c}{} & \multicolumn{3}{c}{Image-to-text Retrieval} &\multicolumn{3}{c}{Text-to-Image Retrieval} \\
        \hline
	     Model & R@1 ($\uparrow$)  & R@5 ($\uparrow$) & R@10 ($\uparrow$)  & R@1 ($\uparrow$) & R@5 ($\uparrow$)  & R@10 ($\uparrow$) \\
        \hline
         ViLT ~\cite{kim2021vilt} & 83.5 & 96.7 & 98.6 & 64.4 & 88.7 & 93.8 \\
         UNITER ~\cite{chen2019uniter} & 87.3 & 98.0 & 99.2 & 75.6 & 94.1 & 96.8 \\
         Frozen ~\cite{bain2021frozen} & - & - & - & 61.0 & 87.5 & 92.7 \\
         ALBEF ~\cite{li2021align} & 94.3 & 99.4  & 99.8 & 82.8  & 96.7 & 98.4 \\
         BLIP ~\cite{li2022blip} & \textbf{96.6} & \textbf{99.8} & 100.0 & 87.2 & 97.5 & 98.8 \\
         ALIGN ~\cite{li2022align} & 95.3 & 99.8 & 100.0 & 84.9 & 97.4 & 98.6 \\
         SINGULARITY ~\cite{lei2022revealing} & 93.3 & 99.4 & 99.8 & 81.4 & 95.8 & 97.9 \\
         BridgeTower ~\cite{xu2023bridgetower} & 94.7 & 99.61 & 100.0 & 85.8 & 97.6 & 98.9 \\
         ManageTower ~\cite{xu-etal-2023-managertower} & 95.6 & - & - &  86.5 & - & - \\
         \hline
         \textbf{MuMUR} & 90.5 & 99.7 & 99.8 & \textbf{92.0} & \textbf{99.5} & \textbf{99.9}  \\
	   \hline
		\end{tabular}}
		\caption{Text-to-image and image-to-text retrieval results on Flickr30k ~\cite{plummer2015flickr30k} dataset split. Recall at rank $1$ (R@1)$\uparrow$, rank $5$ (R@5)$\uparrow$, rank $10$ (R@10)$\uparrow$ are reported. Results of other methods taken from mentioned references. MuMUR significantly outperforms previous models in text-to-image retrieval while achieving comparable results in image-to-text retrieval.}
        \label{tab: Flickr30k}
	\end{center}
\end{table*}

\begin{table}[ht]
	\begin{center}
		\resizebox{\textwidth}{!}
		{
		\begin{tabular}{c|ccccccc} %
        \hline
	    Model & de & cs & zh & ru  & sw & es \\
        \hline
	   m-BERT (zero-shot) &	11.1&	8.2	& 6.9&	7.9	&	1.4 &	12 \\
m-BERT MMP (zero-shot)	&15&	11.2&	8.4&	11&		3.4&	15.1 \\
XLM-R (zero-shot)&	16.3&	16	&14.9	&15.4&	7.7&	17.3 \\
XLM-MMP (zero-shot)&	19.4&	19.3	&18.2&	19.1&		8.4	&20.4 \\
m-BERT (fine-tune)&	18.2&	16.9&	16.2&	16.5&	13	&18.5 \\
XLM- R + MMP (fine-tune)&	21.1 &	20.7 &	20 &	20.5 &		14.4 &	21.9 \\
\hline
	    \textbf{MuMUR - fr (zero-shot)}  & \textbf{27.4} &	\textbf{28.2} &	\textbf{24.1} &	\textbf{26.6} &		\textbf{22.5} &	\textbf{26.5} \\
	    \hline
		\end{tabular}}
		\caption{Text-to-video retrieval (R@1 metric) results on MSR-VTT - multilingual \cite{huang2021multilingual}. Results of other methods taken from \cite{huang2021multilingual}. Our model is trained on Charades dataset and using only french language and evaluated in a zero-shot setting on MSRVTT multilingual dataset. In zero-shot evaluation on other languages, our model significantly outperforms previous models trained in both zero-shot and fine-tuning setting.}
		\label{tab: MSRVTT_multiling}
	\end{center}
\end{table}

\begin{table}[ht]
	\begin{center}
		\resizebox{0.7\textwidth}{!}
		{
		\begin{tabular}{c|ccc} %
        \hline
	    Model & de & fr & cs \\
        \hline
	   EmbN ~\cite{wang2018learning} &	60.3	&54.8	&46.3
\\
PAR. EmbN ~\cite{gella2017image}	&62.6	&60.6&	54.1
 \\
S-LIWE ~\cite{wehrmann2019language} &	72.1	&63.4	&59.4
	\\
MULE ~\cite{kim2020mule} &	64.1	&62.3&	57.7
\\
SMALR ~\cite{burns2020learning} &	69.8	&65.9&	64.8
 \\
M3P ~\cite{ni2021m3p} &	82.1	&67.3&	65
  \\
\hline
	    \textbf{MuMUR}  & \textbf{85.2} &	\textbf{70.5} &	\textbf{69.3} \\
	    \hline
		\end{tabular}}
		\caption{Text-to-image retrieval (R@1 metric) results on Multi30K \cite{elliott2016multi30k} dataset (de - German, fr - French and cs - Czech). Results of other methods are taken from \cite{ni2021m3p}. Our model is fine-tuned for each language and evaluated on the corresponding test data.}
		\label{tab: Multi30k}
	\end{center}
\end{table}

\section{Results and Discussion}
\subsection{Evaluation on English video retrieval datasets}
In Table \ref{tab: MSRVTT_9k} we report the results of our proposed approach on MSVRTT-9k dataset. It can be observed that the difference between CLIP based models and other models is very significant ($> 5\%$). Therefore, it explains the incentive to build our model using CLIP features. On MSRVTT-9k split, our model significantly outperforms CLIP4Clip model on all the metrics in both text-to-video and video-to-text retrieval settings. VCM employs a knowledge graph between video and text modalities making its performance superior to other models in a video-to-text retrieval task. Our model surpasses VCM significantly in all the metrics elucidating that the multilingual representations serve as a powerful knowledge transfer. Moreover, our approach outperforms MCQ and MILES which are pretrained on WebVid-2M data, initialized with CLIP features, employing additional semantic information like parts-of-speech. This validates that our model doesn't require any pretraining on videos and structural knowledge injection. The multilingual text representations in our model effectively serves this purpose. 

In Tables \ref{tab: MSRVTT_7k}, \ref{tab: MSVD}, \ref{tab: DiDeMo} and \ref{tab: Charades} we report the results on MSRVTT-7k, MSVD, DiDeMo and Charades datasets respectively. Our model outperforms all the previous approaches across all the metrics on all the datasets. For the MSRVTT-7k split, our model achieves a significant boost of 2.1\%, 2.5\% and 4.4\% in R$@$1, R$@$5 and R$@$10 respectively compared to the previous baselines. For the MSVD dataset, we notice an improvement of 0.2\%, 0.5\% and 0.3\% in R$@$1, R$@$5 and R$@$10 respectively. MSVD is a relatively smaller dataset with test size of 670 videos and hence, the improvements are relatively marginal.

For the DiDeMo dataset, our model showed a marginal boost of 0.2\% in R$@1$ but a significant boost of 4.3\% and 2.9\% in R$@$5 and R$@$10 respectively compared to the previous approaches. For the Charades dataset, our model outperformed previous approaches by 0.9\% in R$@$1 and by a significant margin of 4.6\%, 7.6\% and 6.0\% in R$@$5, R$@$10 and MedianR respectively. ECLIPSE uses audio as additional information for video retrieval. We showed that multilingual text acts as a better knowledge transfer input.

\subsection{Evaluation on English image retrieval datasets}
Next, we measure the performance of MuMUR on English image retrieval dataset Flickr30k. The results are reported in the table ~\ref{tab: Flickr30k} in both image-to-text and text-to-image settings. As shown in the table, MuMUR achieves comparable performance to previous approaches on image-to-text retrieval. Moreover, we observe that MuMUR significantly outperforms these models by 5.5\% in image-to-text retrieval.
Note that, these models are pretrained on large amount of image-text pairs and fine-tuned on Flickr30k dataset. In contrast, our model uses a small amount of multilingual data and achieves remarkable results in both the settings. This validates that multilingual data acts as superior knowledge transfer even for image retrieval.

\subsection{Evaluation on multilingual video retrieval datasets}
In addition to the monolingual datasets, we also evaluate the proposed approach on multilingual video retrieval datasets. Specifically, we use the model trained only using French captions and test on 6 languages such as German (de), Czech (cs), Chinese (zh), Swahili (sw), Russian (ru) and Spanish (es) in a zero-shot setting. Table \ref{tab: MSRVTT_multiling} shows the results on MSRVTT-multilingual dataset. Our model achieved a significant boost of 8.2\% (average) in R$@$1 in a zero-shot setting. It is worth noting that our model in a zero-short evaluation outperformed the previous approaches fine-tuned on these languages by a huge margin of 6.1\% (average). MMP \cite{huang2021multilingual} is pretrained on the large scale multilingual dataset HowTo100M on 9 languages. However, our model trained on just 1 language outperformed MMP. This shows that our dual cross-modal (DCM) encoder block can effectively learn the association among video, English and multilingual representations even when large video pretraining is not involved.

\subsection{Evaluation on multilingual image retrieval datasets}
We also evaluate the proposed model MuMUR on the multilingual image retrieval dataset Multi30K. Table ~\ref{tab: Multi30k} show the results on 3 languages German (de), French (fr) and Czech (cs) fine-tuned for these languages. As shown in the table, MuMUR significantly outperforms previous models by 3.1\% for German (de), 3.2\% for French (fr) and 4.3\% for Czech (cs) languages. 

\subsection{Ablation studies}
\subsubsection{Effect of multilingual knowledge transfer}
\begin{figure}[ht]
    \centering
    \includegraphics[width=0.8\columnwidth]{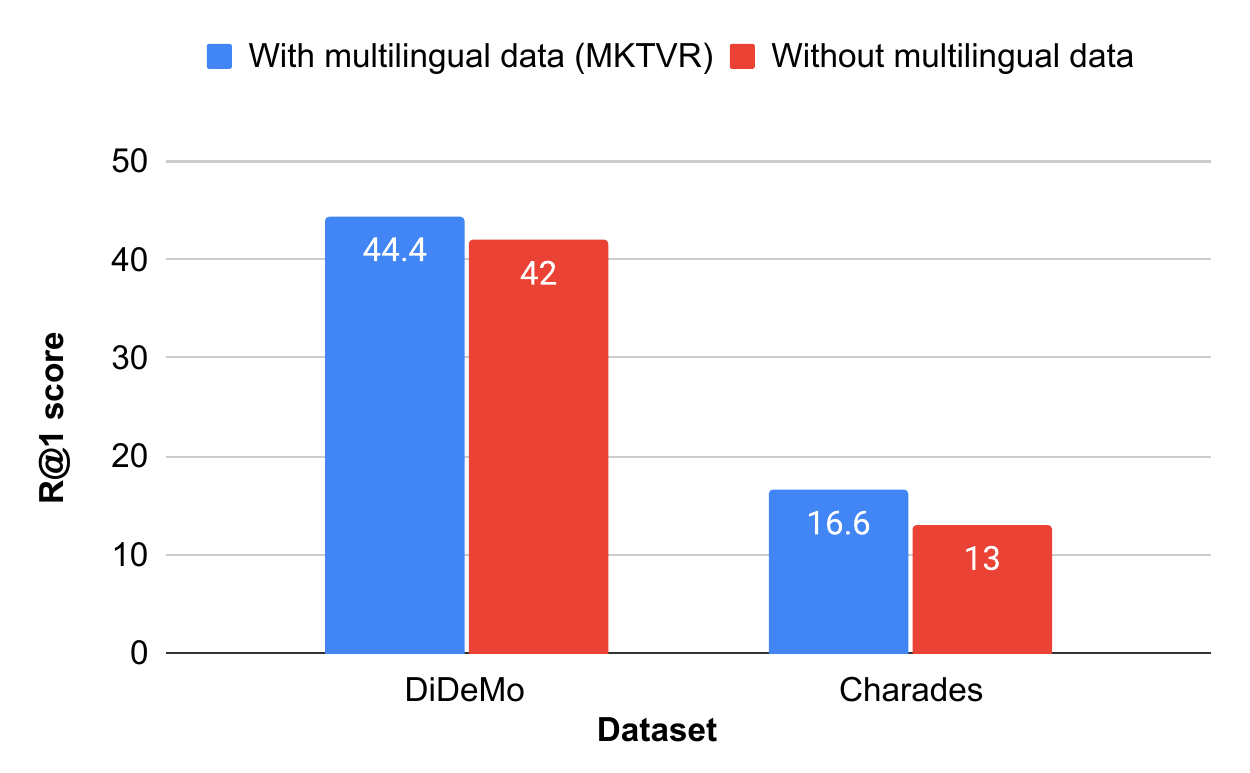}
    \caption{Comparison of models with and without using multilingual data as input. The first model takes as input only video and English text captions whereas the second model takes video, English text and multilingual text captions as input. As shown in the figure, using multilingual text data as knowledge transfer significantly improved the performance.}
    \label{fig:english_multi_full}
\end{figure}

We investigate the effect of multilingual knowledge transfer on the video-retrieval performance. Precisely, we train a model without the multilingual text encoder keeping the rest of the architecture intact. As shown in Figure \ref{fig:english_multi_full}, using multilingual data as knowledge transfer significantly improved the performance on DiDeMo and Charades datasets. The improvement is 2.4\% for DiDeMo and 3.62\% for Charades datasets.

\begin{figure}
    \centering
    \includegraphics[width=0.8\columnwidth]{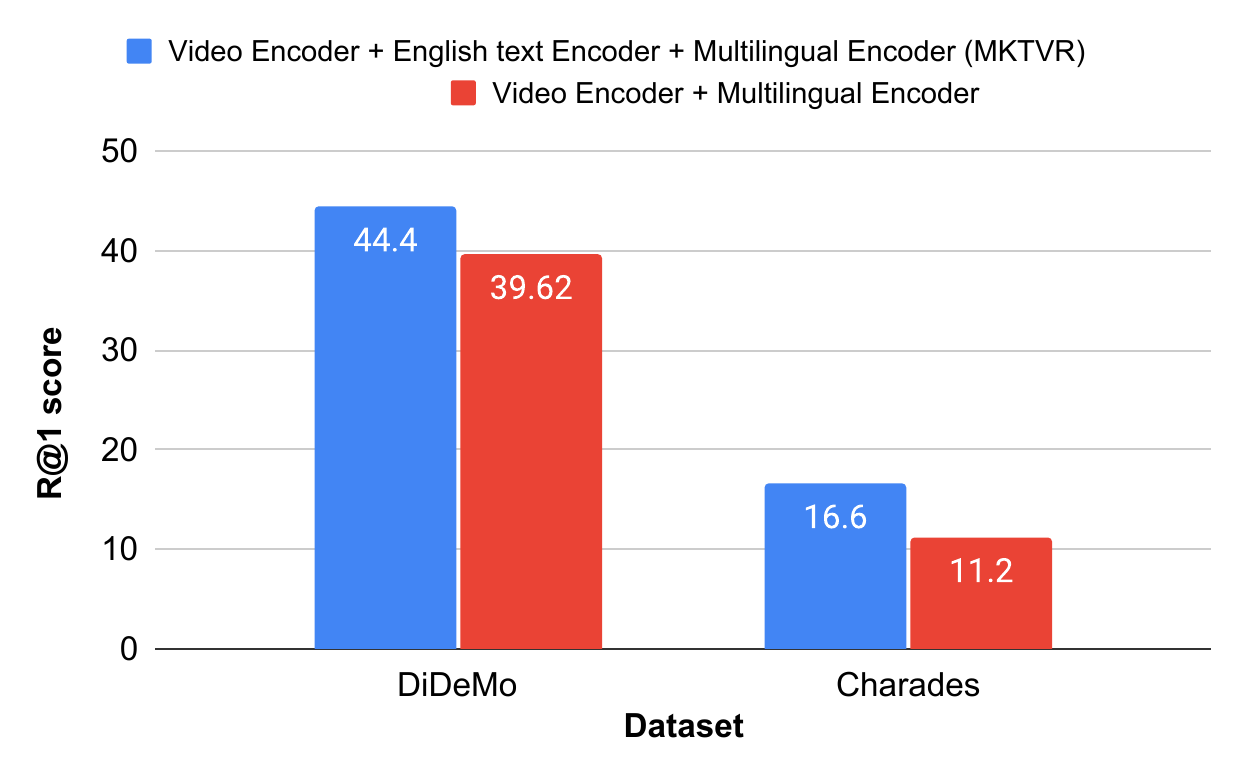}
    \caption{Comparison of models consisting of only multilingual text encoder and multilingual text encoder + English text encoder. Using a separate English text encoder for encoding English text captions outperforms the model using multilingual text encoder to encode English text captions}
    \label{fig:multi_nomulti}
\end{figure}

\begin{figure}
    \centering
    \includegraphics[width=0.8\columnwidth]{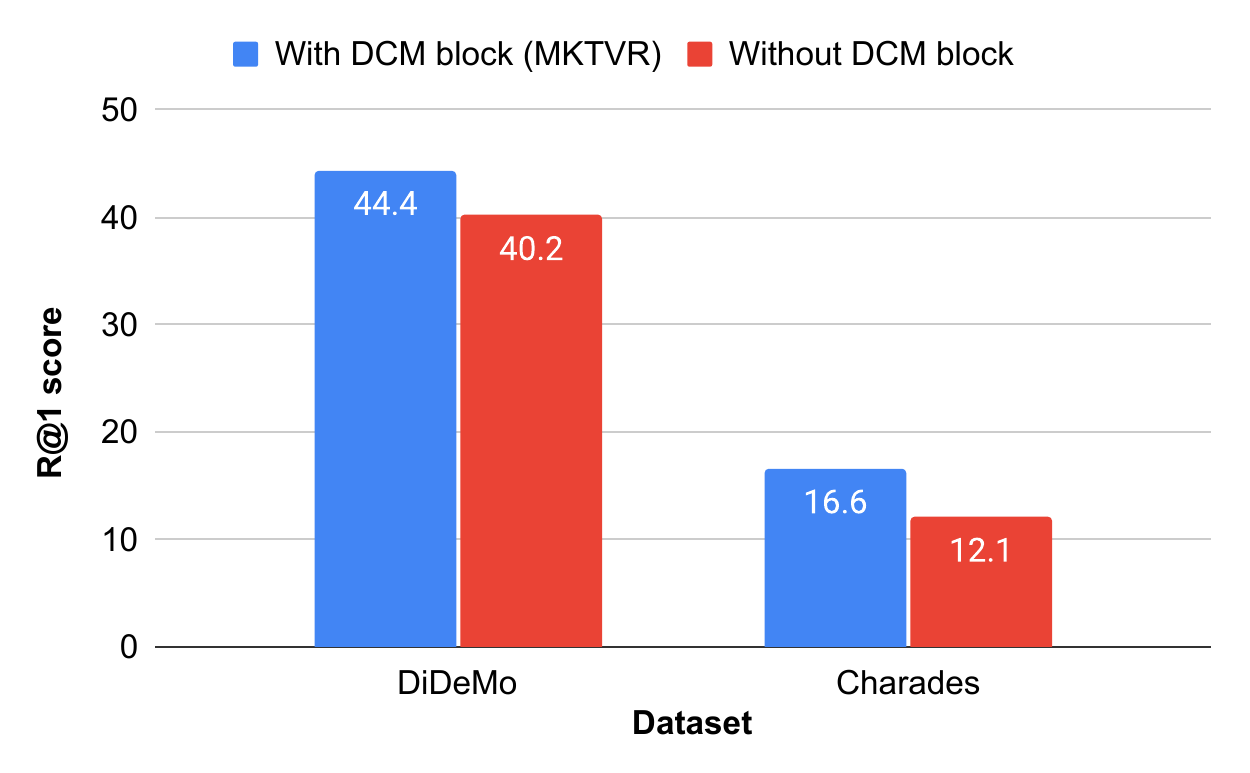}
    \caption{Comparison of models with and without DCM block in the architecture. Using DCM block in the architecture showed superior performance to models without the DCM block.}
    \label{fig:dcm_nodcm}
\end{figure}
\begin{figure}
    \centering
    \includegraphics[width=0.8\columnwidth]{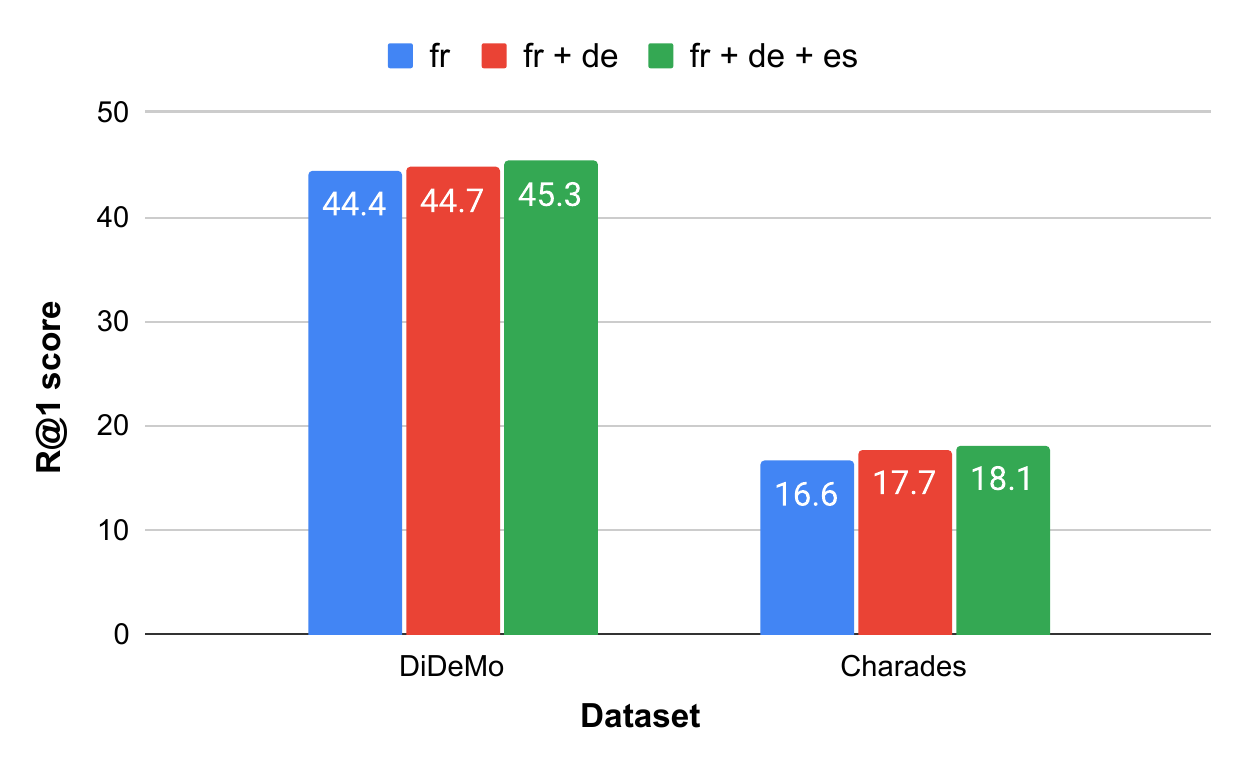}
    \caption{Comparison of MuMUR trained with different multilingual caption data. It is evident from the figure that training with more languages improved the performance.}
    \label{fig:languages}
\end{figure}

\begin{figure}
    \centering
    \includegraphics[width=0.6\columnwidth]{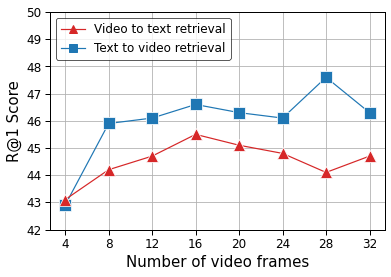}
    \caption{Figure shows the effect of the number of video frames used to train MuMUR model on MSRVTT-9k dataset. Results demonstrate that R@1 score is the highest when 16 frames are used in text-to-video and video-to-text settings.}
    \label{fig:msrvtt_frames_abl}
\end{figure}

\begin{figure}
    \centering
    \includegraphics[width=0.6
    \columnwidth]{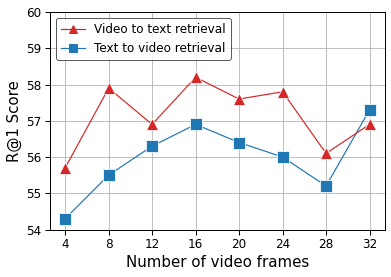}
    \caption{Figure demonstrates the performance of MuMUR for varying number of video frames on MSVD dataset (single caption evaluation). It is evident that the R@1 score is maximum at 16 frames when evaluated in both text-to-video and video-to-text settings.}
    \label{fig:msvd_frames_abl}
\end{figure}

\begin{figure}
    \centering
    \includegraphics[width=0.6\columnwidth]{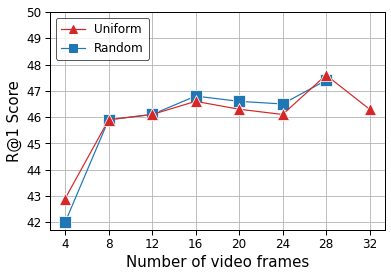}
    \caption{We ablate the sampling strategy used for selecting video frames on MSRVTT-9k dataset. We observe that uniform and random sampling techniques achieve similar performance for all the video frames.}
    \label{fig:msrvtt_sampling_abl}
\end{figure}

\begin{figure}
    \centering
    \includegraphics[width=0.6
    \columnwidth]{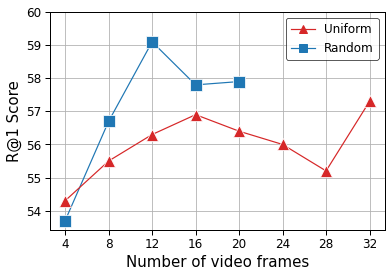}
    \caption{Figure illustrates the comparative performance of random and uniform video sampling techniques on MSVD dataset. It is evident that random significantly outperforms uniform sampling but fails for larger number of video frames.}
    \label{fig:msvd_sampling_abl}
\end{figure}
\subsubsection{Using only multilingual text encoder}
Next, we ablate the choice of using an English text encoder. We validated previously that multilingual data improves the performance of video retrieval. This raises the question: \textit{Why a separate English text encoder is required if multilingual text encoder can be used for both English text and multilingual text representations?} In Figure \ref{fig:multi_nomulti}, we show the results of two different model variants. The first model uses a separate English text encoder to encode English text captions whereas in the second model, both the English text and multilingual text are encoded using the same multilingual text encoder. Results show that encoding English text captions using a separate English text encoder surpasses the model using multilingual text encoder to encode both English text and multilingual text. Multilingual pretraining employs a part of English data whereas the English text encoder is pretrained on a comparatively larger English data. Hence, leveraging a separate English text encoder showed much superior performance to using multilingual text encoder for English text.
\subsubsection{Effectiveness of dual cross encoder block}
Next, we ablate the effectiveness of dual cross encoder block. We train a model without the DCM block and directly compute the loss between video representations and English text representations and video representations and multilingual text representations. From Figure \ref{fig:dcm_nodcm}, we can see that the model using DCM block achieves better performance than the model without the encoder block. This justifies our motivation to use DCM block in our model.
\subsubsection{Training with more languages}
Next, we ablate training our model with more than one language. Concretely we train our model with German (de) and Spanish (es) captions. These languages are chosen because their  performance on XNLI dataset \cite{conneau2018xnli} is comparable to English. The results are shown in Figure \ref{fig:languages} and it is seen that training with more languages improved the performance on video retrieval. These results validate that multilingual data act as an effective knowledge transfer mechanism for improving video retrieval.
\subsubsection{Effect of video frames}
We investigate the impact of the frequency of video frames on the retrieval performance. We train the MuMUR model with increasing number of video frames in intervals of 4. Figures ~\ref{fig:msrvtt_frames_abl} and ~\ref{fig:msvd_frames_abl} demonstrate the results of this ablation study on MSRVTT-9k and MSVD datasets respectively. As shown in the figure, the R@1 score is the highest when the model is trained with 16 video frames in both text-to-video and video-to-text retrieval settings. Therefore, these results motivate the use of 16 frames for training MuMUR on video retrieval datasets.
\subsubsection{Effect of video frame sampling}
Following, we study the impact of choosing video frames at random vs with an uniform manner. Figures ~\ref{fig:msrvtt_sampling_abl} and ~\ref{fig:msvd_sampling_abl} illustrate the results for varying video frames on MSRVTT-9k and MSVD datasets respectively. It is clear from the figures that uniform and random sampling strategies achieve nearly the same performance. However in case of MSVD, we observe that random sampling performed much better compared to uniform sampling. Nevertheless we see that random sampling fails for video frames greater than 20.
\section{Conclusion}
In this paper we introduced MuMUR, a multilingual knowledge transfer framework to improve the performance of multimodal retrieval. We constructed multilingual captions using off-the-shelf state-of-the-art machine translation models. We then proposed a CLIP-based model that enables multilingual knowledge transfer using a dual cross-modal encoder block. Experiment results on six standard multimodal retrieval datasets showed that our framework achieved state-of-the-art results on all the datasets. Finally, our model also showed superior performance to previous approaches on multilingual retrieval datasets in a zero-shot and fine-tune settings. In the future, we will focus on more efficient ways of multilingual knowledge transfer for multimodal retrieval.

\section{Acknowledgment}
We are grateful to the Habana R\&D team, especially Chaitanya Lolla, Sebastian Rogawski, and Radhakrishna Giduthuri, who provided crucial support for execution of this model on Intel Habana Gaudi AI accelerators.

\section{Declarations}
\subsection{Ethical Approval}
This declaration is not applicable.
\subsection{Competing interests}
The authors declare that they do not have competing interests.
\subsection{Authors' contributions}
A.M designed the model, performed major experiments and wrote the first draft. E.A helped in experiments and manuscript writing. G.B.M.S helped in experiments and manuscript writing. S.R helped in experiments and manuscript writing. S-Y.T helped manuscript writing and supervised the project. G.B supervised the project and reviewed the manuscript. V.L supervised the project and reviewed the manuscript.
\subsection{Funding}
This declaration is not applicable.
\subsection{Availability of data and materials}
The datasets used in this work are publicly accessible and code will be made public after paper acceptance.

\bibliographystyle{splncs04}
\bibliography{sn-bibliography}

\end{document}